Running head: [The K-tool topical knowledge test generator]

# Towards an automatic method for generating
# topical vocabulary test forms for specific reading passages

**Michael Flor, Zuowei Wang, Paul Deane, & Tenaha O'Reilly**

ETS, Princeton, NJ

Background knowledge is typically needed for successful comprehension of topical and domain specific reading passages, such as in the STEM domain. However, there are few automated measures of student knowledge that can be readily deployed and scored in time to make predictions on whether a given student will likely be able to understand a specific content area text. In this paper, we present our effort in developing K-tool, an automated system for generating topical vocabulary tests that measure students' background knowledge related to a specific text. The system automatically detects the topic of a given text and produces topical vocabulary items based on their relationship with the topic. This information is used to automatically generate background knowledge forms that contain words that are highly related to the topic and words that share similar features but do not share high associations to the topic. Prior research indicates that performance on such tasks can help determine whether a student is likely to understand a particular text based on their knowledge state. The described system is intended for use with middle and high school student population of native speakers of English. It is designed to handle single reading passages and is not dependent on any corpus or text collection. In this paper, we describe the system architecture and present an initial evaluation of the system outputs.

*Corresponding author:* Michael Flor, E-mail: mflor@ets.org

This research report describes the development of a prototype automated tool for the generation of vocabulary tests from topical reading passages. In section 1, we present the context and motivation for developing this tool. Section 2 presents the tool and the computational linguistic aspects that power its capabilities. Section 3 presents an evaluation study that assessed the acceptability of generated test items. Discussion and limitations are presented in section 4.

## Section 1: Background and Context

### Why measure students' background knowledge in the context of reading?

The importance of background knowledge for successful reading comprehension has been emphasized by reading researchers for a long time (Simonsmeier et al., 2021; Smith et al., 2021; Castles et al., 2018; Ozuru et al., 2009; Cromley & Azevedo, 2007; Kintsch, 2004; Shapiro, 2004). In short, students with more relevant knowledge on a topic tend to have better comprehension than students with less relevant knowledge. While background knowledge plays a key role in reading comprehension, it is rarely assessed when comprehension is measured. We argue that there is potential added value for measuring students' knowledge before they read a text, in either an assessment context or non-assessment context. In an assessment context, a measure of background knowledge could help contextualize a reading score. A low score on a comprehension test may signal difficulties in understanding, or a low level of background knowledge may impede understanding. Having a measure of knowledge could help disentangle these two interpretations and improve reporting as well as support the validity of a reading test.

In a non-assessment context like in a regular classroom setting, a teacher may want to know whether and who may be at risk of not understanding a text that the class is about to cover. Such students could be given pre-reading activities that increase their knowledge such as watching a video on the topic, reading a brief summary to provide them with background information, or providing a list of "must know words" that are central to understanding the text. The aim of this paper is to describe the development of the prototype tool and describe the evaluation (item acceptance rate) for the generated test forms.

## How to measure background knowledge efficiently?

There is a wide range of research on the role of background knowledge on reading comprehension. From a theoretical perspective, the Construction Integration model is a good example of why knowledge is important for comprehension. In the Construction Integration model (Kintsch, 2004), the reader's representation of a text, called the "situation model", crucially depends on integrating background knowledge and the text contents. This is in part because texts are often incomplete and therefore require the reader to draw knowledge-based inferences to fill in gaps when forming a coherent situation model.

Despite the important role of background knowledge on comprehension, it can be challenging to measure efficiently. For instance, assessment of content and background knowledge at school has traditionally relied on propositional knowledge, assessed with factual statements, such as multiple-choice questions. Development of such items often requires expertise in the specific topic and is time-consuming. For example, in the study by Cromley and Azevedo (2007), the researchers exerted considerable effort to analyze the passages in a reading comprehension test and to identify the relevant background knowledge that would be important for students understanding of the content. With advances in technology, natural language processing (NLP) systems can now successfully generate multiple-choice questions automatically from texts (Mulla & Gharpure, 2023; Kurdi et al., 2020). However, on-demand generation of assessments for measuring the relevant background knowledge for a specific passage remains an active research area.

Converging evidence from reading research has emphasized the central role of vocabulary knowledge for reading comprehension, for both native speakers (McKeown et al., 2017; Perfetti & Stafura, 2014; Pearson et al, 2012) and for second language learners (Schmitt et al., 2011). In addition to general vocabulary, researchers have emphasized the importance of topical knowledge (Wang et al., 2021) and discipline-specific or topic-specific vocabulary for comprehension of reading materials at school (Nagy & Townsend, 2012; Fischer & Frey, 2014).

The intersection between background knowledge and vocabulary knowledge provides a good opportunity for efficiently assessing topical knowledge via a vocabulary test. For example, Stahl (2008) described a Vocabulary Recognition Test (VRT) as an experimenter-constructed task for estimating vocabulary recognition in a content area (e.g., a text about insects). In that study, the test form consisted of 25 words; of them 18 words were related to the content of the informational text and 7 words were used as distractors. The task was administered with paper and pencil to second grade students in a school in the USA. The students had to circle only the words related to the topic, as a kind of yes/no task. As noted by Stahl and Bravo (2010), such a test can be used as a pre-reading activity. It also can be used by a teacher to confirm that groups of students have similar levels of prior knowledge of a topic. However, students who are not familiar with the topical vocabulary are at risk of not comprehending the reading materials. A VRT-type of assessment has the potential to enable teachers to identify such students and provide them with vocabulary support before the reading activity.

**Topical vocabulary as an efficient proxy for background knowledge**

There is evidence to suggest that a topical vocabulary test can be used as an efficient estimate of the level of students' background knowledge. For instance, O'Reilly et al., (2019) investigated the use of a similar yes/no test format for students of grades 9 to 12, with reading materials about ecosystems. Students were presented with a list of terms and were asked to indicate whether each word was related or unrelated to the topic of ecology. The test form included 44 words, of them 26 topical words and 18 distractors. Only nine of the topical words were explicitly mentioned in the text. Procurement of other topical words was done manually by a test developer, but it was facilitated with suggestions derived from an NLP-generated database of word associations.

O'Reilly et al., (2019) demonstrated that such a topical vocabulary test was a viable measure of background knowledge and could predict students' reading comprehension success. It was also efficient because the knowledge test only took 5-7 minutes to complete. Moreover, the test helped identify students who lacked topical knowledge to achieve adequate comprehension. Specifically, the experimenters were able to numerically identify a knowledge threshold such that students who scored below the knowledge threshold were not likely to comprehend a subsequent text related to the topic of the knowledge test, as compared to students who scored above the knowledge threshold. In addition, they also found that knowledge (or lack of thereof) of only the six most strongly related topical words provided an indication of whether a student would be above or below the knowledge threshold. The group below threshold had an average accuracy of 64% on those words, while the group above the threshold had average accuracy of 95%. This suggests that a topical vocabulary test can aid teachers in identifying students who may struggle when reading a given passage because of their overall level and unfamiliarity with "must know" words related to the topic of the text. Interestingly, these must-know words not only included words from the passage, but also words that *did not* appear in the passage, supporting the notion that the network of associations is important as is the vocabulary itself. Notably, the test is not only useful for identification, but it may have instructional implications (e.g., in some cases teachers may choose to teach the "must know" words before reading, to help promote better comprehension outcomes).

Given the advantages of a topical vocabulary recognition test, it may serve as a practical and efficient formative assessment tool to be used by teachers. A crucial step in that direction is to conceptualize a tool that can generate such test forms automatically. In this report, we present a first iteration of the K-tool, an automatic item generation (AIG) system that was designed to generate such tests for domain-specific reading passages. Section 2 presents the system design and implementation details. Section 3 describes an initial evaluation of the system, focusing on the acceptability of the generated test items.

## Section 2: The K-tool system

### Test form composition

The K-tool system was designed to take a single text passage as input and generate one or more topical vocabulary test forms for it. In this study, we focused on passages between 400 and 1500 words, lengths typical for high school students to read in one sitting. Our prior research (O'Reilly et al., 2019; Wang et al., 2021) has demonstrated that a topical vocabulary test with about 30-50 items would achieve a reliability of .90 (Cronbach's alpha). As such, the current K-tool system aims to generate 50 items per passage: 14 terms would be in-document topical terms (TID), 14 out-of-document topical terms (TOD), i.e. from an external lexicon, and 22 non-topical terms (NT) would serve as distractors (also obtained from an external lexicon). The quantity of 50 and the 14:14:22 composition were set as initial targets for developing the current system. In the future such settings might be left for the users to choose, and studies can be conducted on finding optimal settings (see subsection 'Further extensions' in the Discussion section). Instructions on a topical vocabulary test form would ask the test taker to indicate whether each term (e.g., galaxy, tape) is related to the given topic (e.g. astronomy). The proportion of correctly marked terms (accepted topical words and rejected non-topical words) would indicate a student's familiarity with the vocabulary of the given topic. A sample text and a corresponding vocabulary test form are shown in Figure 1. Note that the reading passage itself is *not* part of the test form, because the test is intended to assess topical knowledge *prior* to the reading activities.

The discovery was made by the Kepler space telescope, which is on a mission to find Earthlike exoplanets—planets in orbit around stars other than the sun. Kepler-16b is the 21st confirmed planet that Kepler has detected since its launch in March 2009.

Kepler-16b's star system is located between the constellations Cygnus and Lyra, about 200 light-years from Earth. A light-year equals the distance light travels through space in a single year, or about 5.9 trillion miles. The planet is about the size of Saturn, but, because it's gaseous, scientists don't believe it to be habitable.

Although it has two stars, Kepler-16b is probably much colder than Earth because neither star is as powerful as Earth's sun. One star is 69 percent of the mass of the sun. The other is only 20 percent of the mass of the sun. The two stars—together called a binary star—orbit around a common center. They cross paths every 41 days. The planet orbits around both stars every 229 days.

"We have two stars dancing around each other, and in our line of sight, they eclipse each other," says Laurance Doyle, principal investigator for the SETI (Search for Extraterrestrial Intelligence) Institute in Mountain View, Calif. "Then we have this exquisite little pirouette of the planet going around both of them."

☐ means        ☐ classroom            ☐ half inch
☐ galaxy        ☐ solar batteries      ☐ mission
☐ universe      ☐ orbit                ☐ constellations
☐ space         ☐ true lies            ☐ brightness
☐ article       ☐ nerves               ☐ equation
☐ pace          ☐ sky                  ☐ astronomers
☐ telescope     ☐ observatory          ☐ worker
☐ fraction      ☐ satellite            ☐ mars exploration
☐ starlight     ☐ moon rocks           ☐ comet
☐ alien enemies ☐ stars                ☐ teaching practice
☐ suns          ☐ gravity              ☐ rub
☐ consonant     ☐ stretching exercises ☐ similarity
☐ horizon       ☐ liter                ☐ tape
☐ baking cookies ☐ comma               ☐ values
☐ sphere        ☐ particular           ☐ working atmosphere
☐ asteroid      ☐ planets              ☐ mass
☐ gut           ☐ discovery

**Figure 1** A text passage (part shown) and a corresponding topical vocabulary test form. The instruction for this test form would be "Select all of the terms that are related to the topic of astronomy."

At present, we require that all test terms on a form should be nouns or noun-phrases. We focus on nouns since those typically constitute the most prominent topical vocabulary for texts. We leave utilization of verbs and adjectives in testing for future research – this aspect will be informed by further exploring the educational needs. However, this does not mean that we do not extract topical verbs and adjectives from a text; we just do not include them on a generated test form. The system also uses multi-word expressions (MWEs), because nominal MWEs often represent important topical terminology, such as *potential energy* or *greenhouse gas*.

**High level process for creating forms**

With those requirements, the conceptual design of an AIG system is as follows:

Step 1. Identify the major topic of the text.

Step 2. Select all words (and nominal MWEs) in the text which are strongly related to the main topic.

Step 3. Select nouns (and nominal MWEs) from a general English vocabulary, which do not appear in the text, but are strongly related to the major topic.

\* The nouns from steps 2 and 3 would be used to generate the keys for a test form.

Step 4. Select nouns (and nominal MWEs) from a general English vocabulary, which do not appear in the text, and are absolutely not related to the major topic(s) of the text. This set would be used to generate distractors.

Finding topics in texts is a venerable research area in NLP. It has been long dominated by Latent Dirichlet Allocation (LDA) and related approaches (Vayansky & Kumar, 2020), and recently by neural topic modeling approaches (Wu et al., 2024). However, topic modeling has focused on automatically identifying topics in large collections of texts, and those methods are not directly applicable for analysis of a standalone text document. We are interested in identifying the topics in any standalone document, without relying on prespecified taxonomies of topics, and without relying on any collections of documents.

Another relevant and venerable area of NLP research is automated extraction of keywords and keyphrases (Xie et al., 2023). Such approaches can operate on a single text. However, they are often limited to extracting very few representative keywords. Chau et al. (2021) describe an approach to extracting key concepts from a large textbook, utilizing keyphrase-extraction methods, but also relying on the structure of chapters in a textbook. However, keyword extraction is not a suitable approach in our case, because keywords may reflect different subtopics in a text, and keywords do not exhaustively reflect all the topical words in a text. We need to find in a text all the words (types, not tokens) that represent the main topic of the text.

We introduce a new way for topic detection in a single text passage. From topic modeling literature we borrow the convenient definition of a topic as a set of strongly related words. The outline of our approach is as follows: a) we cluster the words of the text by semantic similarity, b) we find the cluster (or set of clusters) that is most representative of the text. The words in those clusters would be the vocabulary of the main topic of the text. To implement those steps, we utilize neural embeddings for both words/phrases and whole texts.

**Preprocessing steps**

The text of the reading passage is preprocessed as follows: POS tagging, MWE detection, vectorization of the whole text and also vectorization of each word or MWE term.

**POS-tagging**. We identify the part-of-speech (POS) for every word in the text. This is used for identifying the nouns, verbs, adjectives, and other major POS types in a text. We utilize the OpenNLP POS-tagger (https://opennlp.apache.org).

**MWE detection**. Given the text passage, we perform noun phrase MWE detection, using an in-house pre-developed lexicon of 68K nominal MWEs (e.g. *space shuttle*, *abdominal fat*).[1]

**Whole-text vectorization**. The reading passage text is embedded into a single vector representation. This vector serves as a representation of the semantic content of the whole text, and is used later for estimating semantic relatedness of various components to the whole document. We use the MiniLM-L6-v2 model from Sentence BERT library (Reimers and Gurevych, 2019). Since BERT models (and Sentence BERT models) have limits on the number of word tokens (text length) that they can encode (e.g. 400 words), we break longer texts into chunks, vectorize every chunk and then combine the resulting vectors by computing the average vector –a single vector for the whole document (Sannigrahi et al., 2023; Iso et al., 2021; Sun & Nenkova, 2019). The document vector is L2 normalized.[2]

**Term vectorization**. We use a precomputed dictionary of static embeddings for a lexicon of 150K English words. The dictionary was developed by using Sentence BERT as an embedder, with the same MiniLM-L6-v2 model. In the same way, we also developed static embedding vectors for our list of 68K MWEs. Using those resources, every word or MWE for a given text is vectorized via simple dictionary lookup. All vectors are normalized with L2 normalization. The need for static a-contextual embeddings is motivated, as we also use the same vectors to check how strongly words from an external list are related to the topic of a given document (see below). Note that the vector for the document and the vectors for all terms (in the document and external) must be from the same vector space model.

## Finding topics in the text

To find topical groups of terms in the text, we perform clustering of the words and MWEs of the document, using cosine similarity between embedding vectors as the clustering similarity measure.

Since we are interested in topical analysis of terms, we utilize nouns, verbs, and adjectives, and filter out certain types of words that are not useful for such analysis. We use POS tags and stoplists to filter out determiners, prepositions, function words, numbers expressed in digits, and also adverbs, proper nouns (names), interrogatives, demonstratives, do/be/have verb forms, and modal verbs. We include detected MWEs as units, but exclude their parts. For example, if '*space shuttle*' is included, its constituent '*shuttle*' is excluded, unless this word appears by itself elsewhere in the text. Only word types are used for clustering, thus a word-form that appears multiple times in text would be included only once. However, for each word-form we do keep track of how many times it appears in the text.

Since the number of topical term groups in any given text is not known a priori and can vary from text to text, we opt to use a clustering approach that does not require prespecifying the number of clusters. The Affinity Propagation clustering (Fleck & Dueck, 2007) works well for our purposes, as it automatically finds the optimal number of clusters for each text. For words (and MWEs) of a given text, we run affinity propagation clustering, using vector cosine similarity as the similarity measure, until convergence for 10 iterations, and collect the resulting term clusters.

For each cluster we compute a centroid vector that will represent that cluster. The centroid is computed as a weighted average of cluster term vectors, where the weights are the counts of the terms in the text (a.k.a. *tf* weighting). The centroid vector is also L2 normalized.

After those steps, we can estimate which term clusters represent the more central topics of the text. We compute cosine similarity between the vector of the whole document, and the centroid vector of each cluster, and sort the clusters by their similarity to the whole document. Clusters that are most similar to the whole document represent the central, most important topics of the document. To illustrate this point, we present clusters that were generated for a 314-word-long informational passage about thunderstorms (Table 1 and Figure 2).

**Table 1** The top seven clusters for terms from a text about thunderstorms,
sorted by cluster cosine similarity to the whole text

| Cluster # | Cluster's cosine to document | Terms |
|---|---|---|
| 1 | 0.531 | thunder, thunderstorms, severe weather, clouds, hail, winds, tornadoes, lightning, meteorologists |
| 2 | 0.347 | condenses, precipitation, water, droplets, moisture, moist air |
| 3 | 0.325 | atmosphere, conditions, atmospheric conditions, weather, air masses |
| 4 | 0.242 | featuring, unstable, collide, meet, events, causes, occur, cumulonimbus |
| 5 | 0.192 | warm air, air, sun, heating, air temperature, warm, freeze, cooler air |
| 6 | 0.163 | elevated, carry, lift, lifting, rises, movement, force |
| 7 | 0.162 | surface, terrain, mountains |

## Creating sufficient vocabulary for multiple test forms

Our current design for a single topical vocabulary test-form requires 50 terms (all of them nouns or nominal MWEs), of them 14 topical terms from the text, 14 topical terms from external lexicon, and 22 topically unrelated terms as distractors. These requirements are set for a *single* test form. However, in some cases, it might be beneficial to generate more than one test form. For example, we may want to have both an easy and more difficult form to accommodate larger variations in the level of student knowledge (the rationale is presented below, in the section 'Generating multiple test forms'). To allow for the generation of multiple test forms, we need to oversample the vocabulary.

With this in mind, to begin generating a test form for a text, we need to select the required number of topical terms from the top clusters that were obtained. To allow for more flexibility, we introduce a more general term-selection process, which allows for generating multiple test-forms for a given text. The idea is to over-select terms into a pool of accepted terms from which multiple test forms can be generated. The process has three steps: selecting topical terms from the text, topical terms from the external general vocabulary, and the distractor terms.

We generate a pool of topical terms from the text, by over-selecting nominal terms from the top topical clusters (usually the three top clusters). By default, we select 25 nominal terms from those clusters. The number obviously needs to be higher than the 14 TID terms for a form. A good quota might be 28, but for some texts 28 TID terms are not always available; so we set the initial quota to 25. This setting is likely to be modified in the future (see subsection 'Further extensions' in the Discussion section).

The next step is to select topical terms from outside of the given text, and to select non-topical terms to be used as distractors. For this we use a general list of English nouns (about 30K lemmas, a subset of the 150K general list), and the list of MWEs (about 68K entries), as the supply lists from which candidates are drawn. For both of those lists we have corresponding embedding vectors prearranged as general precomputed resources that are used by the K-tool.

Selecting terms for an out-of-document topical list amounts to: a) scanning the vocabulary supply lists, b) filtering out any terms that already appear in the text, and c) selecting words that have sufficient semantic similarity to the in-document topical terms. For example, if a document has the words '*car*', '*road*' and '*drive*' among its top topical terms, we may wish to bring from the external list such words as '*driver*', '*passenger*', '*motor*', '*traffic*', etc.

The notion of 'sufficient semantic similarity' requires some elaboration. One approach could be to find a certain threshold value of similarity to be used as a cutoff. However, we opted for a more dynamic approach. We sort the out-of-document terms by their similarity to the major topics of the document and pick the *n* top rated terms as needed. For this we define a pool of top topical-in-document terms, which is simply the list of all terms from the three top term-clusters of the document. An external candidate term needs to be semantically similar or semantically related to this pool of terms.

The measure for such computation could be simply the cosine similarity between vectors of candidate terms (external list) and the vectors of the in-document topical terms, or through word co-occurrences. Both approaches are well known in the computational linguistic literature. In this study, we combined the cosine similarity measure with the co-occurrence measure. A first-order cooccurrence-based approach reflects the notion that words that frequently occur together are topically related (Schütze & Pedersen 1997). The second-order distributional similarity approach reflects the notion that words occurring within similar contexts are semantically similar or related (Lin 1998; Turney & Pantel, 2010; Bullinaria & Levy, 2012). Although some studies compared the two approaches (Liebeskind et al., 2018; Purandare & Pedersen, 2004), their combination has also been explored (Flor et al., 2019). Vector cosine helps emphasize semantic similarity ('car' and 'truck'), while co-occurrence helps emphasize semantic relatedness ('car' and 'wheel').

In the current work, for each candidate vocabulary term we compute its support in the document as follows:

$$Support(T_c) = \sum_{j=0}^{n} \Big( \big( cosine(T_c, T_{dj}) + PNPMI(T_c, T_{dj}) \big) \times log_{10} \big( count(T_{dj}) + 1 \big) \Big)$$

where $T_c$ is a new candidate term, $T_{dj}$ is the *j*-th word from the list of *N* top in-document topical terms, and *count($T_{dj}$)* is the number of occurrences of $T_{dj}$ in the document.[3] PNPMI is Positive Normalized Pointwise Mutual Information, which is a variant of the well-known PMI measure. Pointwise Mutual Information is defined as follows (Church and Hanks, 1990):

$$PMI = log_2 \frac{p(a,b)}{p(a) \times p(b)}$$

where *a* and *b* are words, *p(a)* and *p(b)* are probabilities of each word, *p(a,b)* is the probability of joint occurrence. The probabilities are estimated from counts in a very large corpus. We used a language model of word co-occurrence within paragraphs, trained on a corpus of more than 2 billion words (Flor and Beigman Klebanov, 2014). Normalized PMI (Bouma, 2009) has values constrained in the range (-1, 1), and is defined as:

$$NPMI = \frac{log_2 \dfrac{p(a,b)}{p(a) \times p(b)}}{-log_2 (p(a,b))}$$

PNPMI takes the value of NPMI, or zero if NPMI is negative or if the value for the co-occurrence of words *a* and *b* is not available in the database. When the terms *a* or *b* (or both) are MWEs, we compute their association as average PNPMI value between the words of the first term and the words of the second term.

**Generating multiple test forms**

As stated above, our design allows for generating more than one test form per text. At this stage in system development, we decided to differentiate the generated test forms by their collective difficulty of vocabulary terms. During the term selection process, for each term we retrieve its grade-level estimator (a decimal value) as a measure of its difficulty. Thus, each term in our pool of selected topical terms also has a grade-level value. Then, during form assembly, we select from this pool the terms with the lowest grade level values, and thus generate a relatively 'easy' form, or the terms with the highest grade level values, and thus generate a relatively 'difficult' form. Grade-level estimators for single words come from the VXGL resource, a list of 126K English words mapped to their estimated grade levels (Flor et al., 2024). For multi-word expressions we do not have a predefined resource yet; thus, we estimate the grade level as follows: find the grade level for each constituent word, pick the word with the highest value, increase that value by 25% and use the result as the grade level estimate for the phrase.[4]

Another consideration for form assembly is that the distribution of grade-levels among the topical terms selected from outside-of-the-document (TOD), and also among the distractor non-topical terms (NT), should be approximately equivalent as for the topical-in-document terms (TID). Otherwise, many of the terms on a form may stand out as being too easy or too difficult relative to the other terms. For example, we want to avoid a case where most of the distractors are 'too easy' (very familiar words). To alleviate this, we use the distribution of grade-levels for the topical-in-document terms as our guiding distribution. As noted above, when selecting terms from the external lexicon, we can create rather large pools of suitable candidates (more than just 14 and 22 respectively). We then perform a second round of selection from those 'large pools', ensuring that the distribution of grade levels for terms in those pools approximately matches the distribution of grade levels in the TID pool. Then, during form generation of a test form, we can pick the terms with highest (or lowest) grade-level values from each of the TID, TOD and NT pools, knowing that the grade level distributions in them are already matched. A sample text and two corresponding auto-generated vocabulary test-forms are shown in Figure 2.

The rationale for generation of different test forms is related to the overall estimation of passage and test difficulty. Passage text complexity can be estimated with a variety of methods, such as readability formulas (e.g. Flesch Kincaid), or using a specialized text complexity tool, such as TextEvaluator (Sheehan et al., 2014). However, the difficulty of a test form that is generated for a given passage is not the same as the overall difficulty of the passage. A vocabulary test form is just a collection of nouns (and nominal MWEs), and many of them are not even from the passage. Also, the core function of the K-Tool item generator is to ensure *topical relations* of the included terms. Without additional control measures, the terms, especially those brought from the general lexicon, can vary in their complexity and grade-level suitability. For instance, the non-topical words could be some common everyday words, or they could be quite uncommon words (but still non-topical). Also, given a topic, the topically related words can have considerable variability with respect to their difficulty/familiarity. For example, consider the topic 'cars', with topical words: *engine* (easy) vs. *crankshaft* (hard). Thus, it might be a useful feature to indicate the difficulty of the vocabulary items (as a supplement to the test form). In addition, it also might be desirable to have some control over the difficulty of the overall test form (e.g., relative to the original reading passage). Such flexibility might also be useful when the same passage is used in different grade levels.

Difficulty of the test form can be manipulated, in part, by controlling the estimated difficulty of the terms included on the form (while still keeping the topicality distinctions as needed). Vocabulary difficulty can be estimated by word frequency, or, in our case, by the VXGL resource. Since we can control the estimated difficulty for the form, the next question is then what difficulty we want to impose. A natural 'anchor' can be the range of difficulties for the topical words already included in the passage. This is a 'data-driven' approach – it depends just on the text. Given such a range, we can then add topical and non-topical terms from the lexicon, with estimated grade-levels closer to the 'easy/hard' points of the range. Thus, we can vary the estimated difficulty of the test form relative to the range of grade-levels of the topical words in the text. The test form can be made relatively easy or relatively difficult. We believe that the ability to control the estimated difficulty of a vocabulary test-form provides a flexibility feature for the prototype tool. Whether this feature will be considered useful by users of the K-tool is an open question for future research. It's also an open question whether the predictive accuracy of the test may vary by the predicted difficulty of the form. For example, a very easy form might be less predictive of student comprehension. This aspect would be investigated in future research.

Thunderstorms occur due to the atmospheric conditions that lead to the development of cumulonimbus clouds, which are large and vertically towering clouds associated with heavy precipitation, thunder, lightning, and sometimes severe weather. Several key factors contribute to the formation of thunderstorms:

1. Moisture: Warm and moist air near the Earth's surface is a crucial component. As the warm air rises, it cools and condenses into water droplets, forming clouds.

2. Instability: The atmosphere needs to be unstable, meaning that the air temperature decreases rapidly with height. This allows the warm, moist air at the surface to rise easily and form updrafts.

3. Lift: Some force is required to lift the warm, moist air. This lift can be provided by several mechanisms, including:
   * Convection: Heating at the Earth's surface (e.g., from the sun) causes air to rise.
   * Fronts: The lifting of air masses along a front where two different air masses meet.
   * Orographic lift: Air is forced to rise over elevated terrain like mountains.

4. Updrafts and Downdrafts: Once the warm, moist air rises and cools, it forms a cumulonimbus cloud. Within these clouds, strong updrafts and downdrafts develop. The updrafts carry water droplets upward, allowing them to freeze and collide, creating electrical charges that lead to lightning. The downdrafts bring cooler air back down to the surface.

5. Condensation and Precipitation: As the air rises and cools within the cumulonimbus cloud, water droplets combine and grow larger, eventually falling as precipitation. The rapid movement of air and water within the cloud contributes to the development of an electrical charge, leading to lightning and thunder.

The combination of these factors creates the dynamic and often intense weather associated with thunderstorms. While many thunderstorms are harmless, some can become severe, featuring strong winds, hail, and tornadoes. Understanding the conditions that lead to thunderstorm formation helps meteorologists predict and monitor these weather events.

a.

b.

| | | |
|---|---|---|
| ☐ costs | ☐ atmosphere | ☐ heavy fog |
| ☐ smog | ☐ separation | ☐ tape |
| ☐ majors | ☐ post | ☐ storm drains |
| ☐ monkey | ☐ demonstration | ☐ rubbing |
| ☐ thunder | ☐ gloom | ☐ light breeze |
| ☐ clouds | ☐ values | ☐ sad fact |
| ☐ snowstorm | ☐ thumb | ☐ score |
| ☐ goat | ☐ rain showers | ☐ quote |
| ☐ precipitation | ☐ idea | ☐ meteorologists |
| ☐ hail | ☐ winds | ☐ fire |
| ☐ dust jacket | ☐ sleet | ☐ thunderstorms |
| ☐ rule | ☐ tornadoes | ☐ moisture |
| ☐ fly | ☐ yes | ☐ lightning |
| ☐ raindrop | ☐ reward | ☐ cumulonimbus |
| ☐ cold drizzle | ☐ weather | ☐ lever |
| ☐ gain | ☐ draw | ☐ rainstorm |
| ☐ classroom | ☐ air | |

c.

| | | |
|---|---|---|
| ☐ discount coupons | ☐ turbulence | ☐ troposphere |
| ☐ aerosol | ☐ moist air | ☐ cyclone |
| ☐ meteorologists | ☐ ionosphere | ☐ betterment |
| ☐ airflow | ☐ tornadoes | ☐ apprenticeship |
| ☐ moisture | ☐ recitation | ☐ lightning |
| ☐ inclination | ☐ specification | ☐ paradigm |
| ☐ cumulonimbus | ☐ airwave | ☐ downdrafts |
| ☐ precipitation | ☐ retrospective review | ☐ droplets |
| ☐ participation | ☐ typhoon | ☐ ozone |
| ☐ airborne particles | ☐ qualification | ☐ portfolio |
| ☐ severe weather | ☐ income fund | ☐ updrafts |
| ☐ atmosphere | ☐ thunderstorms | ☐ vortex |
| ☐ stratosphere | ☐ resolution | ☐ antiquity |
| ☐ undergraduate | ☐ courtship behavior | ☐ locomotion |
| ☐ atmospheric conditions | ☐ symbiosis | ☐ overcast skies |
| ☐ conception | ☐ professorship | ☐ vapor pressure |
| ☐ microcomputer | ☐ creditor | |

**Figure 2** A text passage about thunderstorms (a),

and two generated topical vocabulary test-forms: easy (b) and difficult (c).

The instruction for these test forms could be "Select all of the terms that are related to the topic of weather conditions."

## Section 3: Evaluation

This section describes an initial evaluation of the K-tool system. The purpose of this evaluation was to assess the acceptability of items that are generated by the system. This is known as 'intrinsic evaluation' – it is concerned only with adequacy of the outputs and not its use or efficacy in the classroom. An intrinsic evaluation is necessary to confirm that we have a viable prototype system prior to any potential future field tests that might be conducted with teachers and schools.

### Method

Twenty reading passages were selected from ReadWorks.org, a nonprofit organization that provides school reading materials for K-12 grades in the USA. All twenty passages were expository texts oriented on STEM topics for grades 9-10 (physics, life sciences, technology), and ranged in length from 485 to 1465 words (average 1058 words). We focused on STEM because such passages are typically topical and usually have specifically topical vocabulary, and also because STEM passages usually introduce topics that are not widely familiar to K-12 students – the kind of passages for which assessment of prior knowledge may be very relevant. We chose high-school level texts since this level is the most complex in the K-12 range.

For each passage, two vocabulary test forms were generated, of 'easy' and 'hard' relative difficulty respectively. All vocabulary terms are nouns or noun-phrases.

It turns out that for our texts it was not always possible to produce 14 different TID terms (nouns) for the easy test form, and 14 other TID terms for the difficult test form. As it can happen, the number of distinct topical nouns in a text, especially a shorter text, can be less than 28. In such cases, the system can still produce two forms per text, but the forms will share some of the topical terms. However, we also found that for four out of twenty texts, the K-tool system could not find even 14 in-document topical nouns; for those texts, it found 9, 10, 11, and 13 terms respectively.

Due to scarcity of TID terms, we have the following complication. We would expect to obtain 2000 terms in total: 50 terms per form, with 2 forms per document and 20 documents. However, we have only 1971 total terms for evaluation. Out of those, 140 terms were shared in easy and difficult test forms, so we have 1831 unique term-document cases for evaluation. (See more on scarcity in the 'Technical Limitations' section below).

Our evaluation was concerned with acceptability of each term on its test form. It was carried out by two evaluators. One evaluator was an intern, a college senior studying psychology; the second evaluator was one of the authors, an experienced linguist. The protocol of annotation was as follows. Given a reading passage and a test form automatically generated for the text, the human evaluator's task was to annotate each term on the form. The annotation for each term was a binary decision: yes/no acceptable. All three types of terms, *topical-in-document*, *topical-out-of-document*, and *non-topical*, were clearly identified as such beside each term (those labels were automatically provided by the system). The criteria for acceptability were as follows. Topical terms have to be clearly related to the topic of the passage. Non-topical terms have to be clearly unrelated to the topic. The text passage was available to the raters during annotation.

The annotation task was defined by two of the authors. Prior to the main task, the raters conducted a training session (with two passages) and then a pilot annotation (on one passage), and achieved an agreement of 94%.

## Results

Of the 1831 unique terms, evaluators agreed in 1766 cases; that is agreement of 96.45%. Cohen's Kappa is 0.7495 (substantial agreement, according to Landis and Koch, 1977). For individual texts, agreement was relatively high, ranging from 92% to 99%. For the easy forms, agreement ranged between 90% and 100% across the 20 texts; for the difficult forms, agreement ranged between 89% and 100%.

Next, we consider term acceptability. We used a strict criterion of acceptance: a term is accepted only if *both* evaluators accepted it. With such strict criterion, the term acceptance rate was 1658 of 1831, that is 90.55%. Thus, a large majority of the automatically generated test-terms were considered adequate for vocabulary testing purposes.

Table 2 provides a breakdown of the accepted terms by the term *type* and *source*. The acceptance of non-topical terms was high, 97%. The algorithm seems capable of supplying non-topical distractors for the vocabulary test. The system's ability to extract topical vocabulary from texts was even higher, 99.5%. However, the ability to supply topical terms from an external lexicon was less accurate, with an acceptance rate of 74.5%. A breakdown for each passage is given in Appendix 1.

**Table 2** Acceptance rates for evaluated terms.

| Term type | Accepted | Total | Acceptance rate |
|---|---|---|---|
| NT: Non-topical (distractors) | 852 | 880 | 0.970 |
| TID: Topical in-document | 389 | 391 | 0.995 |
| TOD: Topical from lexicon | 417 | 560 | 0.745 |
| **Total** | 1658 | 1831 | 0.906 |

We set to investigate the system performance for topical-out-of-document (TOD) terms. Figure 3 presents the acceptance rates for TOD terms in the test forms generated for 20 texts. For each text we have 28 TOD terms, so the percents are relative to that max number. Seven texts have acceptance rates above 0.8. One text in particular stands out with a very low acceptance rate (0.39). That text is about scientists investigating sediment rocks that provide evidence about oxygen in the atmosphere on Earth millions of years ago. The topical terms from the document (TID) are a mix of atmosphere-related terms and chemistry-related terms, such as *air, waters, iron, powder, weathers, oxygen, mineral, crust, atmosphere, reactions, layers, compounds, sediments, ozone*. The TOD terms include terms that are related to the topic via the chemistry theme but are not directly relevant to the discussion in the text. Among the TOD terms rejected by evaluators are *fuel combustion, carbon monoxide poisoning, photosynthesis, gasoline, aerosol, ammonia*. What we encounter in this case is that the topic of the text is at an *intersection of two domains* (atmosphere and chemistry), and that TOD words that are strongly related to just *one* of the domains might be too 'far' from the relevant thematic intersection. Another text in our collection discusses biological aspects of the connection between dinosaurs and modern birds. The strict-criterion acceptance rate for TOD terms for that text was 0.64. Among the evaluator-rejected TOD terms for that text we find (*gastropod, cephalopod, mammal, mammoth, aquatic plants)*. It is easy to see the general biological relation, but evaluators considered whose terms to be too far from the specific topic of the text.

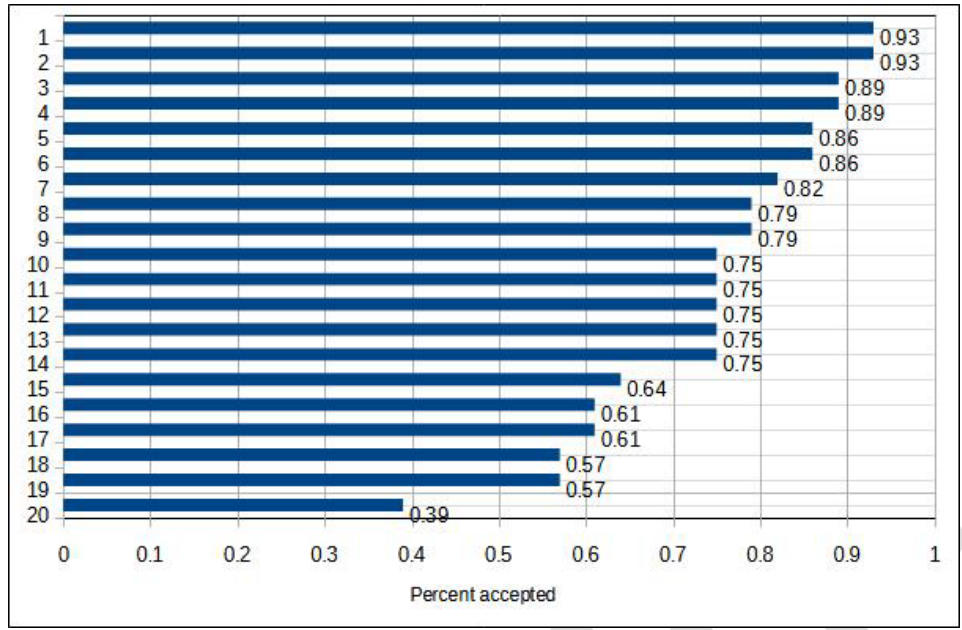

**Figure 3** Acceptance rates for topical-out-of-document terms, for 20 texts.

Such analysis suggests that we may need to reconsider some of our initial assumptions. The topical vocabulary test is intended to test students' knowledge from two perspectives: vocabulary in a *given content area*, and vocabulary related to a *specific reading passage*. But what is the relevant content area? If we define the content area too broadly (e.g. evolutionary biology), such definition might be too general for a specific text. On the other hand, if we define the relevant content area too narrowly, we might be in a 'tight corner' and may have difficulty finding enough relevant terms for a test. A discussion between the annotators revealed that they sometimes tended to focus on the relevance of a candidate TOD term to the actual text, rather than to a broader content area.

Moreover, we also need to consider cases when texts intersect domains that are not usually discussed together. For example, one of the texts in our collection discussed sound phenomena that are encountered in a baseball game. The terminology was a mix of baseball-related terms and terms related to the physics of sound waves. The acceptance rate of TOD terms in this case was 0.57. This example illustrates the need to consider not only the scope and grain size of a 'content area' (e.g., sound waves, ear anatomy, sound location,), but also what is the main content area itself – is it baseball or is it physics?  If one sticks to the 'obvious', assuming it is both baseball and physics, then many TOD terms are rejected because they are associated with either baseball or physics, but not both. Such examples illustrate that we need to investigate further what kinds of 'content area' or 'domain knowledge' we can measure with vocabulary tests. The issue is not vocabulary, but rather what should be the relevant content area and how do we demarcate it for assessment purposes.

## Section 4: Discussion

Students' reading comprehension difficulties can originate from a variety of different sources. For example, our prior work has demonstrated that comprehension problems could be associated with students' poor decoding skills (Wang et al., 2019; Wang et al., 2020; Wang et al., 2024), being unfamiliar with the reading topic (O'Reilly et al., 2019; Wang et al., 2021), or inadequate reading fluency (Sabatini et al., 2019). Identifying the bottleneck of comprehension can inform instructional activities. Relevant to the current study, some studies demonstrated that students' topical vocabulary is a good indicator of potential comprehension difficulties. O'Reilly et al. (2019) showed that student performance on a five-minute topical vocabulary test, administered before students read materials on the topic, was a good indicator for whether students could achieve adequate comprehension when reading. Further, students' topical knowledge can also serve as an indicator for the cultural relevance of the reading materials (Wang et al., in press).

We envision that a topical vocabulary test can be an important component in reading instruction. A low score on a traditional text-comprehension test usually indicates that there is a problem, but it does not indicate what are the sources of the problem. Measuring background knowledge allows one to identify one of the major potential problems for low comprehension. In such context, the K-tool is aimed at generating topical vocabulary tests, which could be administered before the actual reading. Such tests can be quick to administer, providing instructors with an opportunity to identify potential knowledge gaps before students get engaged in extended reading. As a result, instructors can address the knowledge gaps either by pre-teaching, using other relevant interventions, or maybe even changing the reading assignment. In this way, assessment of background knowledge can help teachers to adapt instruction to student needs and help students build understanding before covering a text in class.

Our approach in designing the K-tool stemmed from two core notions. First, a topical vocabulary test may become a useful instructional component, as outlined above. Second, we believe that such a tool should give teachers maximum freedom in selecting the reading passages. The teacher brings the passage they want to use in assigned reading, and the tool would generate the tests. That is why the tool includes automated detection of the major topic of the passage – to allow for a wide variety of passages, without prescribing supported topics in advance.

One limitation of the current system is that it is oriented for reading passages that have domain-specific topical terminology. Expository passages with STEM contents are a prominent example of such texts. Other types of reading materials can also involve special background knowledge with a prominent terminological aspect. For example, texts about specific sports, e.g. American football (Wang et al. 2021). Our initial evaluation of the K-Tool was on STEM texts. We did not yet test the tool on news or social sciences texts, where distinct topical vocabulary is less prominent. Additional research would be needed to investigate whether the current computational approach can be extended to support texts with less prominent topical vocabularies. Another possible avenue for research is to have the tool compute automatically whether a given submitted text has as strong topical vocabulary 'signature'. With such a component, the tool could automatically flag texts that are suitable or unsuitable for topical vocabulary testing.

**Technical limitations**

In this section, we note two technical limitations of our current system that can be immediately addressed as first steps in further development: a) handling of passage titles, and b) generating the topic label for the test form.

Many reading passages used in educational settings come with titles, and in many cases the titles can be indicative of the passage main topic. However, in some cases, the passage titles are not topic-indicative (e.g. catchy or humorous phrases, literary allusions, etc.). One of the obvious steps in further research would be to integrate passage titles in the topic detection process.

Another aspect of our work is yet incomplete – the provision of the topic label for the test form. When the topical vocabulary-test form is composed, it needs to have a topic statement, such as "Which of the following terms is related to the topic of <TopicLabel>?". For example. O'Reilly et al. (2019) presented a test form that asked: "Which of the following terms is related to the topic of ecology?" We intend for our k-tool system to generate such topic-labels automatically, together with the list of test-terms, for any given document.

Automatic provision of topic labels has been explored in computational linguistics literature. Bhatia et al. (2016) used word embeddings to select topic labels for topics derived from LDA processing (LDA topics are technically just sets of words). They utilized Wikipedia article titles as candidates from which topic labels could be obtained, and selected the title that had the best aggregate cosine similarity with each of the words for a given topical set. The situation with k-tool is more complicated, due to several constraints. First, sometimes the suitable topic-label occurs in the document and might be listed among the topical terms. Such is the case for the passage about thunderstorms (Figure 2), where the word 'thunderstorms' appears in the passage itself. In principle, it could be chosen as the query label (for posing the question) and removed from the list of the terms on the test-form. Yet, there is another constraint – a query label needs to be general enough, and not any specific term can be used. For example, for the thunderstorms passage, we might not want to have 'hail' or 'wind' selected as the query topic. The query topical label needs to be general and abstract enough to subsume all the related topical terms under it. Another requirement for the topical query label is that it needs to be easily understood by the students. For example, 'plate tectonics' might be a suitable general term for a text passage, but it might be unfamiliar to students (of a given grade level). Then it would make little sense to ask students which terms are related to a topic whose label they don't understand ('earthquakes' might be more suitable is such case). We consider that an automatic system could provide several candidate query labels, to allow the teacher/researcher to choose the most appropriate one. Thus, this aspect could be addressed in future development.

**Further extensions**

In this section we describe potential extensions beyond the scope of the current prototype system. Those include: a) using terms beyond nouns and noun phrases, b) considerations of text length, and c) considerations regarding the number of items on a test form.

The current K-tool system generates test forms only with nouns and noun phrases. Nouns and noun phrases carry the most topical significance in a text. However, some verbs and adjectives can be distinctively topical. It is yet an open question to what extent verbs and adjectives should be integrated into topical vocabulary tests. In addition, potential inclusion of verbs and adjectives as topical terms holds the promise to alleviate scarcity of in-document topical terms which we have encountered with some texts.

Another limitation of the current study is text length. The system showed promising generation results for reading passages of length from about 400 to 1500 words. Passages shorter than 400 words may lack the sufficient number of topical words, or they may have just enough topical words to support only one test form. One way to address this in the future is to consider shorter test forms, requiring less topical terms from the document. Passages longer than 1500 words might have enough topical terms, but they may involve more themes. The crucial aspect of such passages might not be length per se, but their lexical diversity (how many different topical terms are used). Those aspects can be investigated in further research.

While the different levels of form difficulty are estimated via grade-level mappings, the empirical difficulty of test forms needs to be investigated in a field study with actual student populations. In addition, we need to consider the length of test-forms. For the composition of a test form, we currently require 50 terms, with a breakdown of 14 in-document topical terms, 14 out-of-document topical terms, and 22 distractors. For comparison, O'Reilly et al. (2019) used test forms with 44 items, of them 26 topical and 18 distractors; Wang et al. (2021) used forms with 30 items (in both those studies items were written by experts and tests administered to high school students). Stahl & Bravo (2010) tested students in the second grade, using forms with 25 items. Technically, we can change the required number of items/terms per form, even allow users to control this setting, but there are open questions. Is there an optimal number of items per form? Does it depend on the grade level of the students? Maybe forms need to be longer for higher grades and shorter for lower grades. Since topical vocabulary testing is not yet a widespread educational practice, additional research will be needed to investigate those aspects, and how they influence the accuracy of the vocabulary recognition test to predict students' reading comprehension. However, having an automated system for generation of topical vocabulary forms can ease the generation of such tests and thus contribute to the adoption of such tests in research and in educational practice,

**Considerations for future research**

For the current K-tool implementation we used association and similarity measures to estimate semantic relatedness between terms. While such measures provide estimates on continuous scales, for using them in the K-tool we needed to set some thresholds. The topical vocabulary test uses a yes/no response for each item. Thus, each word/term must be either obviously related to the topic or obviously unrelated. Any item that is somewhere 'in-between' might be considered as inconsistent (debatable) for scoring. Our major concern was that items must be scorable without much dispute. The evaluation results indicate that this requirement was met to a large extent. However, there might be some potential to use varying levels of semantic relatedness to make items more or less difficult. For example, non-topical items that are 'strongly unrelated' to the topic may make a test form easier, while non-topical items that are 'less unrelated' to the topic might make the test more difficult. Varying the levels of semantic relatedness for the topical terms can also potentially influence test form difficulty. Thus, varying the levels of semantic relatedness between terms and topics might be a good direction for future research.

## Conclusions

We presented a prototype computational system, K-tool, designed for automatic generation of vocabulary tests to evaluate whether students have the necessary background knowledge to understand content-specific reading materials. As a proof of concept, K-tool was evaluated on texts for US grades 9-10. A high proportion of terms generated by the system were deemed acceptable by human evaluators. The encouraging results call for further research and development, including collecting empirical data for validation and obtaining teacher usability feedback for refinement.

Once fully established, the system will generate tests on-demand, tailored to the specific content of a selected reading passage. It integrates certain lexical resources but is not dependent on any collection of educational texts or any taxonomy of domains or topics. As such, it may become a useful tool for teachers, and possibly for reading-content developers. Moreover, K-tool has an integrated capability to generate test forms of different levels of difficulty by using grade-level mappings of terms.

The K-tool is generally intended for teachers' informal use in class, as an aid to teaching, and not as a formal assessment tool. The test forms generated by K-tool can be printed for use in class or can be integrated in a computer-based delivery system. They come with associated information that allows scoring student responses automatically. The whole process of administering and scoring such forms can be very quick, and thus provide a way for fast and efficient estimation of students' background knowledge for reading comprehension. Students can take the topical test before a reading activity.

It should be noted that this research is the first step in a larger agenda, and many aspects of this system require additional investigation. For example, more studies need to be conducted to establish validity evidence for use of such system in schools.

## Acknowledgments

This paper describes a new method and a tool for automatically generating items and assembling forms for assessments of background knowledge based upon information extracted from individual texts. The presently implemented system uses some previously developed resources. One of the resources utilized int this tool, specifically the lexicon of multi-word expressions, is based in part on foundational research funded by the U.S. Department of Education Institute of Education Sciences though Grant R305A080647. Previous research on background knowledge and reading comprehension was conducted with funding from the U.S. Department of Education Institute of Education Sciences though Grants R305A150176; R305F100005. The opinions expressed are those of the authors and do not represent views of the IES. The specific ideas and designs for the fully automated tool described in this paper were not part of any previous research.

**Notes**

1 The 68K list of MWEs is a subset of an even larger set of MWEs that was developed at ETS by Paul Deane and Bob Krovetz in the past. That work is yet unpublished, though it is rooted in prior work, Deane (2005) and Deane & Krovetz (2015), that used statistical approaches over large language corpora to extract lists of MWEs. For the current work, we used only the noun MWEs.

2 See https://en.wikipedia.org/wiki/Cosine_similarity#L2-normalized_Euclidean_distance

3 Our current approach is a simple linear combination of cooccurrence (PNPMI) and vector similarity (cosine), weighted proportionally to the (log of) frequency of occurrence of the anchor topical terms from the document. In principle, it is possible to introduce a coefficient of importance $\alpha$ (e.g. $\alpha \cdot \text{cosine} + (1-\alpha) \cdot \text{PNPMI}$ ), and learn the optimal value of $\alpha$ from empirical data. We leave such exploration for future work.

4 The general idea is that the grade-level of an expression should be higher than that of the highest individual word in it. The 25% is just a plausible heuristic. Estimating the complexity (grade-level) of MWEs is not a developed area yet.

# Appendix

## Acceptance rates for generated terms

Table A1 below presents the final acceptance rates of terms for each of the 20 passages, with a breakdown by type of term. Note that acceptance was defined with a strict criterion, meaning the term was accepted only if both annotators marked it as accepted. The categories below are TID (topical terms from inside the document), TOD (topical terms retrieved from out of the document – from the lexicon), and NT (non-topical terms, i.e. distractors, retrieved from the lexicon). The column labelled 'Words' indicates the passage length (word count). The column GL provides the grade-level estimation for the passage (Flesh Kincaid Grade Level). The column "Domain" presents the general content domain of the passage.

**Table A1.** Acceptance rates for evaluated terms, for each passage

| Text ID | Words | GL | Domain | TID | TOD | NT |
|---------|-------|------|-------------------|------|------|------|
| AP | 886 | 9.2 | Astronomy | 100% | 79% | 100% |
| DB | 1000 | 10.1 | Biology | 100% | 64% | 98% |
| EE | 804 | 9.7 | Physics | 91% | 82% | 100% |
| FF | 1282 | 10.7 | Biology | 100% | 94% | 98% |
| HM | 1185 | 9.8 | Biology | 100% | 89% | 100% |
| OG | 717 | 7.7 | Astronomy | 100% | 79% | 100% |
| NSS | 557 | 11.5 | Ecology | 100% | 75% | 86% |
| SYF | 1161 | 13.3 | Biology | 100% | 93% | 100% |
| OEA | 1229 | 9.8 | Atmosphere | 100% | 39% | 95% |
| PC | 601 | 9 | Physics & Sport | 100% | 75% | 100% |
| RWK | 1154 | 9.9 | Physics | 100% | 75% | 98% |
| SC | 827 | 9.9 | Technology | 100% | 75% | 98% |
| SME | 1310 | 13.1 | Biology | 100% | 93% | 89% |
| SAT | 1143 | 11.9 | Astronomy | 100% | 89% | 93% |
| BBSS | 484 | 7.7 | Physics & Sport | 100% | 61% | 95% |
| SPW | 1057 | 9.7 | Physics & Sport | 100% | 57% | 100% |

| SOB | 789 | 7.8 | Physics & Sport | 100% | 57% | 98% |
|-----|-----|-----|-----------------|------|-----|-----|
| SOL | 1194 | 10.8 | Biology | 100% | 86% | 98% |
| WHC | 710 | 9.4 | Geography | 100% | 75% | 95% |
| CFE | 1448 | 10.3 | Ecology | 100% | 61% | 100% |
| Macro average across texts | | | | 99.5% | 75.4% | 97% |
| Micro average across all terms | | | | 99.5% | 74.5% | 97% |

**Suggested citation:**

Flor, M., Wang, Z., Deane, P., & O'Reilly, T. (2025). *Automatic generation of topical vocabulary test forms for reading comprehension* (Research Report No. RR-25-…). ETS. https://doi.org/...

**Action Editor:** xxx

**Reviewers:** xxx and xxx

Find other ETS-published reports by searching the ETS ReSEARCHER database.